\documentclass[runningheads]{llncs}
\usepackage{subcaption}
\usepackage{graphicx}
\usepackage{todonotes}
\usepackage{svg}
\usepackage{bm}
\usepackage{soul}
\begin{document}

\title{Disentangling Shape and Pose for Object-Centric Deep Active Inference Models}
\titlerunning{Disentangling Shape and Pose}

% Uncomment for camera ready if accepted
\author{Stefano Ferraro \and
 Toon Van de Maele \and
 Pietro Mazzaglia \and \\
 Tim Verbelen \and 
 Bart Dhoedt}%

\institute{IDLab, Department of Information Technology \\
 Ghent University - imec \\ 
 Ghent, Belgium \\
 \email{stefano.ferraro@ugent.be}
}

\authorrunning{S. Ferraro et al.}
\maketitle              % typeset the header of the contribution
\begin{abstract}
Active inference is a first principles approach for understanding the brain in particular, and sentient agents in general, with the single imperative of minimizing free energy. As such, it provides a computational account for modelling artificial intelligent agents, by defining the agent's generative model and inferring the model parameters, actions and hidden state beliefs. However, the exact specification of the generative model and the hidden state space structure is left to the experimenter, whose design choices influence the resulting behaviour of the agent. Recently, deep learning methods have been proposed to learn a hidden state space structure purely from data, alleviating the experimenter from this tedious design task, but resulting in an entangled, non-interpreteable state space. In this paper, we hypothesize that such a learnt, entangled state space does not necessarily yield the best model in terms of free energy, and that enforcing different factors in the state space can yield a lower model complexity. In particular, we consider the problem of 3D object representation, and focus on different instances of the ShapeNet dataset. We propose a model that factorizes object shape, pose and category, while still learning a representation for each factor using a deep neural network. We show that models, with best disentanglement properties, perform best when adopted by an active agent in reaching preferred observations.

\keywords{Active Inference  \and Object Perception \and Deep Learning \and Disentanglement.}
\end{abstract}

\section{Introduction}

In our daily lives, we manipulate and interact with hundreds of objects without even thinking. In doing so, we make inferences about an object's identity, location in space, 3D structure, look and feel. In short, we learn a generative model of how objects come about~\cite{parr_generative_2021}. Robots however still lack this kind of intuition, and struggle to consistently manipulate a wide variety of objects~\cite{manipulation}. Therefore, in this work, we focus on building object-centric generative models to equip robots with the ability to reason about shape and pose of different object categories, and generalize to novel instances of these categories.

Active inference offers a first principles approach for learning and acting using a generative model, by minimizing (expected) free energy. Recently, deep learning techniques were proposed to learn such generative models from high dimensional sensor data~\cite{catal_learning_2020,Fountas2020DAIMC,Sancaktar2020PixelAI}, which paves the way to more complex application areas such as robot perception~\cite{Lanillos2021AIFRobotics}. In particular, Van de Maele et al.~\cite{van_de_maele_disentangling_2021,vandemaele_2022} introduced object-centric, deep active inference models that enable an agent to infer the pose and identity of a particular object instance. However, this model was restricted to identify unique object instances, i.e. ``this sugar box versus that particular tomato soup can'', instead of more general object categories, i.e. ``mugs versus bottles''. This severely limits generalization, as it requires to learn a novel model for each particular object instance, i.e. for each particular mug.

In this paper, we further extend upon this line of work, by learning object-centric models not by object instance, but by object category. This allows the agent to reduce the number of required object-centric models, as well as to generalize to novel instances of known object categories. Of course, this requires the agent to not only infer object pose and identity, but also the different shapes that comprise this category. An important research question is then how to define and factorize the generative model, i.e. do we need to explicitly split the different latent factors in our model (i.e. shape and pose), or can a latent structure be learnt purely from data, and to what extent is this learnt latent structure factorized?

In the brain, there is also evidence for disentangled representations. For instance, processing visual inputs in primates consists of two pathways: the ventral or ``what'' pathway, which is involved with object identification and recognition, and the dorsal or ``where'' pathway, which processes an object's spatial location~\cite{Mishkin1983}. Similarly, Hawkins et al.  hypothesize that cortical columns in the neocortex represent an object model, capturing their pose in a local reference frame, encoded by cortical grid cells~\cite{hawkins_theory_2017}. This fuels the idea of treating object pose as a first class citizen when learning an object-centric generative model.

In this paper, we present a novel method for learning object-centric models for distinct object categories, that promotes a disentangled representation for shape and pose. We demonstrate how such models can be used for inferring actions that move an agent towards a preferred observation. We show that a better pose-shape disentanglement indeed seems to improve performance, yet further research in this direction is required. In the remainder of the paper we first give an overview on related work, after which we present our method. We present some results on object categories of the ShapeNet database~\cite{shapenet2015}, and conclude the paper with a thorough discussion.

\section{Related work}
\label{sect:sota} 

\textbf{Object-centric models.} Many techniques have been proposed for representing 3D objects using deep neural networks, working with 2D renders \cite{generate_cnn}, 3D voxel representations \cite{3dgan}, point clouds \cite{point_cloud_gen} or implicit signed distance function representations \cite{park_deepsdf_2019,mescheder_occupancy_2019,mildenhall_nerf_2020,sitzmann_siren_2020}. However, none of these take ``action'' into account, i.e. there is no agent that can pick its next viewpoint.

\noindent \textbf{Disentangled representations.} Disentangling the hidden factors of variation of a dataset is an long sought feature for representation learning \cite{representation_learning}. This can be encouraged during training by restricting the capacity of the information bottleneck \cite{bVAE}, by penalizing the total correlation of the latent variables \cite{factorvae,tcvae}, or by matching moments of a factorized prior \cite{dipvae}. It has been shown that disentangled representations yield better performance on down-stream tasks, enabling quicker learning using fewer examples \cite{disentangled}.
\hfill \smallbreak
\noindent \textbf{Deep active inference.} Parameterizing generative models using deep neural networks for active inference has been coined ``deep active inference'' \cite{kai2018}. This enables active inference applications on high-dimensional observations such as pixel inputs~\cite{catal_learning_2020,Fountas2020DAIMC,Sancaktar2020PixelAI}. In this paper, we propose a novel model which encourages a disentangled latent space, and we compare with other deep active inference models such as \cite{catal_learning_2020} and \cite{van_de_maele_active_2021}. For a more extensive review, see \cite{mazzaglia_entropy}.

\section{Object-Centric Deep Active Inference Models}
\label{sect:method} 

In active inference, an agent acts and learns in order to minimize an upper
bound on the negative log evidence of its observations, given its generative model
of the world i.e. the free energy. In this section, we first formally introduce
the different generative models considered for our agents for representing 3D objects. Next we discuss how we instantiate and train these generative models using deep neural networks, and how we encourage the model to disentangle shape and pose.
%Finally, we show how action selection is driven by minimizing expected free energy in the future.

\begin{figure}[t!]
    \centering
    \includegraphics[width=\textwidth]{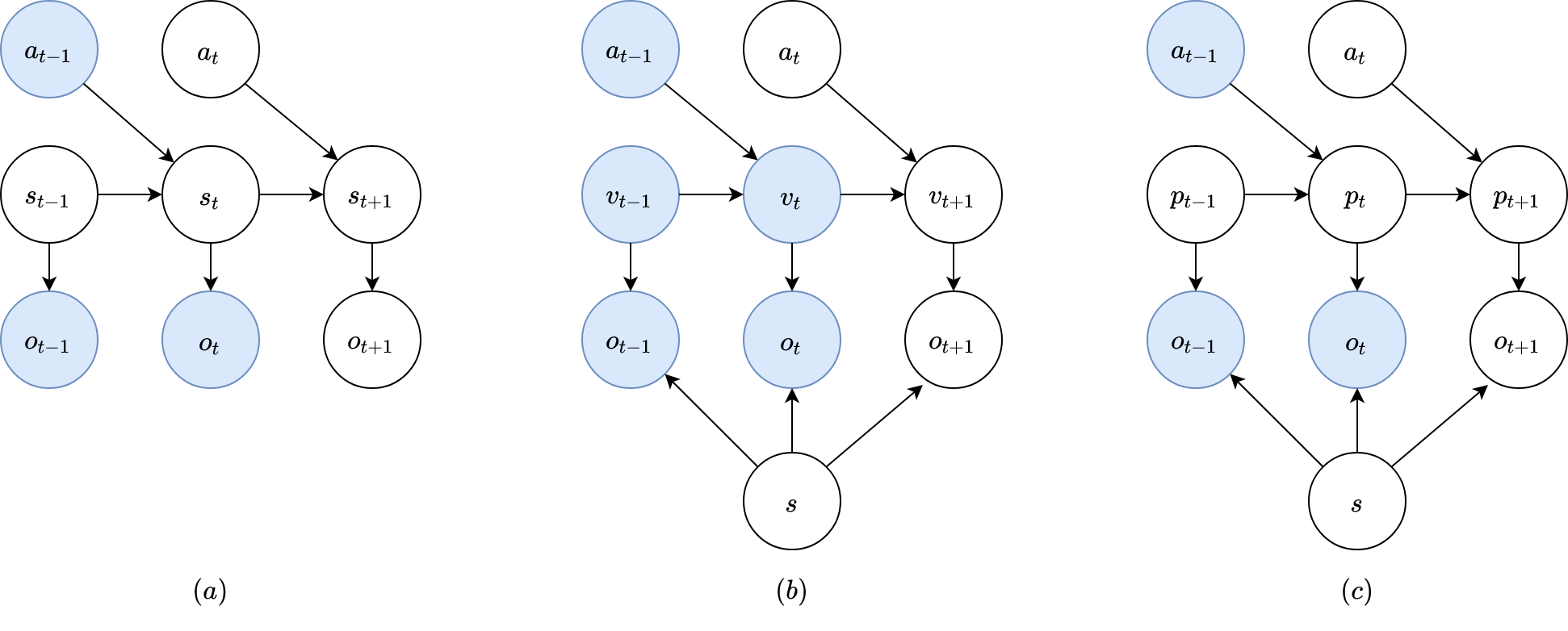}
    \caption{Different generative models for object-centric representations, blue nodes are observed. (a) A generic POMDP model with a hidden state $s_t$ that is transitioned through actions and which generates the observations. (b) The hidden state $s$ encodes the appearance of the object, while actions transition the camera viewpoint $v$ which is assumed to be observable. (c) Similar as (b), but without access to the camera viewpoint, which in this case has to be inferred as a separate pose latent variable $p_t$.}
    \label{fig:models}
\end{figure}

\hfill \break
\noindent \textbf{Generative model.} We consider the same setup as \cite{vandemaele_2022}, in which an agent receives pixel observations $o$ of a 3D object rendered from a certain camera viewpoint $v$, and as an action $a$ can move the camera to a novel viewpoint. The action space is restricted to viewpoints that look at the object, such that the object is always in the center of the observation. 

Figure~\ref{fig:models} depicts different possible choices of generative model to equip the agent with. The first (1a) considers a generic partially observable Markov decision process (POMDP), in which a hidden state $s_t$ encodes all information at timestep $t$ to generate observation $o_t$. Action $a_t$  determines together with the current state $s_t$ how the model transitions to a new state $s_{t+1}$. This a model can be implemented as a variational autoencoder (VAE)~\cite{rezende_stochastic_2014,kingma_auto-encoding_2014}, as shown in \cite{catal_learning_2020,Fountas2020DAIMC}. A second option (1b) is to exploit the environment setup, and assume we can also observe the camera viewpoint $v_t$. Now the agent needs to infer the object shape $s$ which stays fixed over time. This resembles the architecture of a generative query network (GQN), which is trained to predict novel viewpoints of a given a scene \cite{eslami_neural_2018,van_de_maele_active_2021}. Finally, in (1c), we propose our model, in which we have the same structure as (1b), but without access to the ground truth viewpoint. In this case, the model needs to learn a hidden latent representation of the object pose in view $p_t$. This also allows the model to learn a different pose representation than a 3D pose in SO(3), which might be more suited. We call this model a VAEsp, as it is trained in similar vein as (1a), but with a disentangled shape and pose latent.
% TODO : should we refer to appendix for the free energy formula and loss function?

\begin{figure}[t!]
  \centering
  \includegraphics[width=\textwidth]{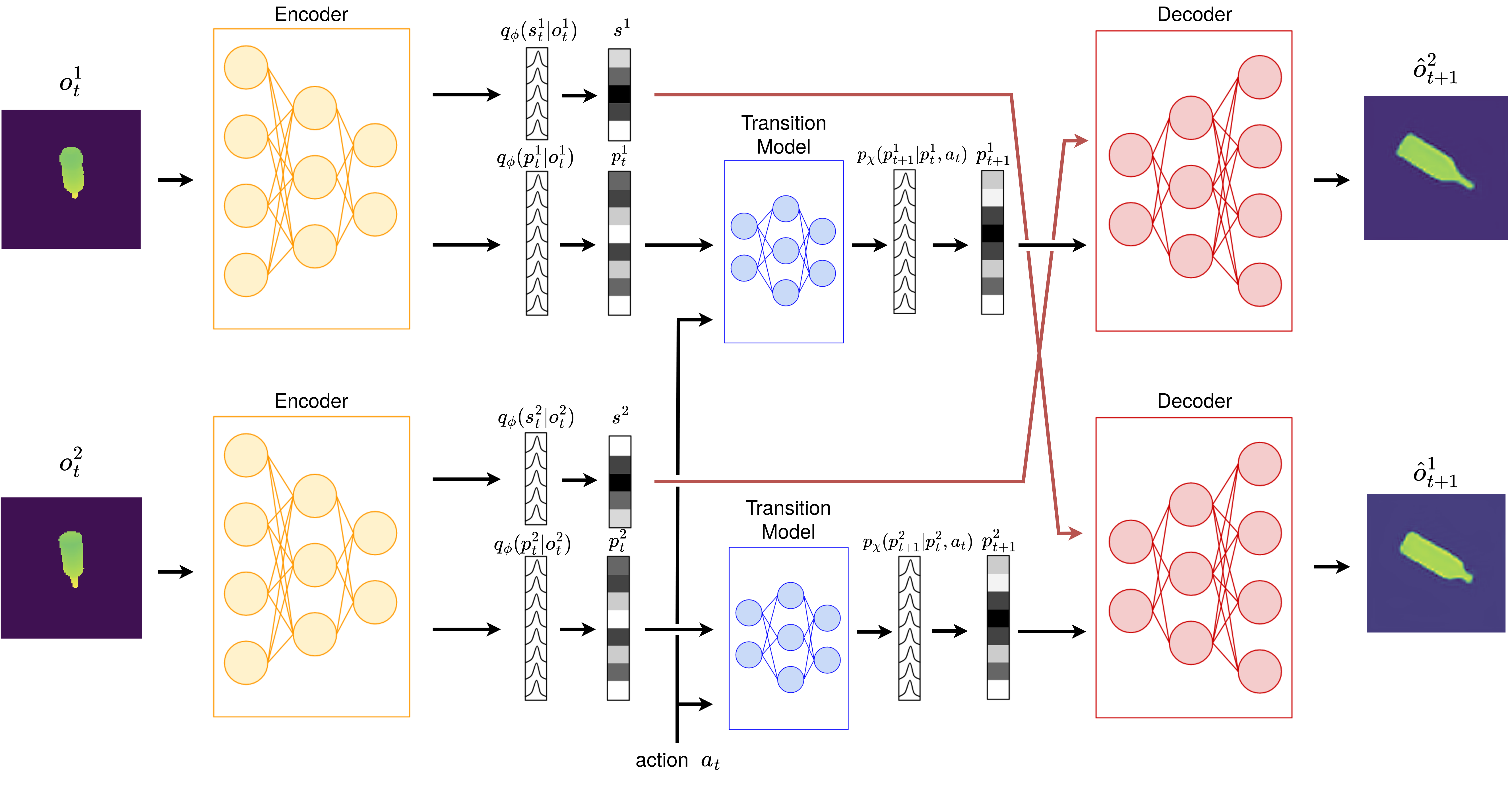}
  \caption{The proposed VAEsp architecture consists of three deep neural networks: an encoder $q_{\phi}$, a transition model $p_\chi$, and a decoder $p_\psi$. By swapping the shape latent samples, we enforce the model to disentangle shape and pose during training.}
  \label{fig:VAEsp}
\end{figure}

\hfill \break
\noindent \textbf{VAEsp.} Our model is parameterized by three deep neural networks: an encoder $q_{\phi}$, a transition model $p_\chi$, and a decoder $p_\psi$, as shown in Figure~\ref{fig:VAEsp}. Observations $o^i$ of object instance $i$ are processed by the encoder $q_\phi$, that outputs a belief over a pose latent $q_{\phi}(p^i_t|o^i_t)$ and a shape latent $q_{\phi}(s^i_t|o^i_t)$. From the pose distribution a sample $p^i_t$ is drawn and fed to the transition model $p_\chi$, paired with an action $a_t$. The output is a belief $p_\chi(p^i_{t+1} | p^i_t, a_t)$. From the transitioned belief a sample $p^i_{t+1}$ is again drawn which is paired with a shape latent sample $s^i$ and input to the decoder $p_\psi(o^i_t | p^i_t, s^i)$. The output of the decoding process is again an image $\hat{o}^i_{t+1}$. These models are jointly trained end-to-end by minimizing free energy, or equivalently, maximizing the evidence lower bound~\cite{vandemaele_2022}. More details on the model architecture and training hyperparameters can be found in Appendix \ref{appendix:details}.

\hfill \break
\textbf{Enforcing disentanglement.} In order to encourage the model to encode object shape features in the shape latent, while encoding object pose in the pose latent, we only offer the pose latent $p_t$ as input to the transition model, whereas the decoder uses both the shape and pose. Similar to~\cite{Huang_2018_ECCV}, in order to further disentangle, we randomly swap the shape latent code for two object instances at train time while keeping the same latent pose, refer to Figure~\ref{fig:VAEsp}.

\section{Experiments}

We train our model on a subset of the ShapeNet dataset~\cite{shapenet2015}. In particular, we use renders of 15 instances of the `mug', `bottle',`bowl' and `can' categories, train a separate model for each category, and evaluate on unseen object instances. We compare our VAEsp approach against a VAE model \cite{catal_learning_2020} that has equal amount of latent dimensions, but without a shape and pose split, and a GQN-like model \cite{van_de_maele_active_2021}, which has access to the ground truth camera viewpoint.

We evaluate the performance of the three considered generative models. First we look at the reconstruction and prediction quality of the models for unseen object instances. Next we investigate how good an agent can move the camera to match a preferred observation by minimizing expected free energy. Finally, we investigate the disentanglement of the resulting latent space.

\begin{table}[t!]
    \centering
    \caption{One-step prediction errors, averaged over the entire test set. MSE (lower the better) and SSIM (higher the better) are considered. }
    \begin{tabular}{c l|c|c|c|c}
    & & bottle & bowl & can & mug \\
    \hline 
    MSE $\Downarrow$ & GQN & $0.473 \pm 0.0874$	& $0.487	\pm 0.141$	& $0.707 \pm 0.1029$	 & $0.656 \pm 0.0918$ \\
    \textbf{} & VAE & ${0.471 \pm 0.0824}$ & $0.486 \pm 0.1487$	& ${0.693 \pm 0.1103}$	& $0.646 \pm	0.0886$ \\
    \textbf{} & VAEsp & $0.480 \pm 0.0879$ & ${0.485 \pm 0.1486}$ &	$0.702 \pm 0.1108$ &	${0.626	\pm 0.0915}$ \\
    
    & & & & & \\

    SSIM $\Uparrow$ & GQN & $0.748 \pm	0.0428$ & $0.814 \pm 0.0233$ &	$0.868 \pm 0.0203$ &	$0.824 \pm 0.0279$\\
    \textbf{} & VAE & $0.828 \pm 0.0238$	& $\mathbf{0.907 \pm 0.0178}$ &	$0.844 \pm 0.0361$ & $\mathbf{0.874 \pm 0.0323}$ \\
    \textbf{} & VAEsp & $\mathbf{0.854 \pm 0.0190}$ &	$0.902 \pm 0.0291$ &	$\mathbf{0.880 \pm 0.0176}$ &	$0.814 \pm 0.0348$ \\
    \end{tabular}
    \label{tab:prediction}
\end{table}

\hfill \break
\noindent \textbf{One-step Prediction.} First, we evaluate all models on prediction quality over a test set of 500 observations of unseen objects in unseen poses. We provide each model with an initial observation which is encoded into a latent state. Next we sample a random action, predict the next latent state using the transition model, for which we reconstruct the observation and compare with a ground truth. We report both pixel-wise mean squared error (MSE) and structural similarity (SSIM) \cite{ssim} in Table~\ref{tab:prediction}. In terms of MSE results are comparable for all the proposed architectures. In terms of SSIM however, VAEsp shows better performance for `bottle' and `can' category. Performance for `bowl' category are comparable to the best performing VAE model. For the `mug' category, the negative gap over the VAE model is consistent. Qualitative results for all models are shown in Appendix~\ref{appendix:qual}.

\begin{table}[b!]
    \centering
    \caption{MSE for the reached pose through the minimization of expected free energy. For each category, 50 meshes are evaluated, where for each object a random pose is sampled from a different object as preferred pose, and the agent should reach this pose.}
    \begin{tabular}{l|c|c|c|c}
    & bottle & bowl & can & mug \\
    \hline 
    GQN & $0.0833 \pm 0.0580$ & $0.0888 \pm 0.0594$ & $0.0806 \pm 0.0547$ & $0.1250 \pm 0.0681$ \\
    VAE & $0.0698 \pm 0.0564$ & $\mathbf{0.0795 \pm 0.0599}$ & $0.0608 \pm 0.0560$ & $0.1247 \pm 0.0656$ \\
    VAEsp & $\mathbf{0.0557 \pm 0.0404}$ & $0.0799 \pm 0.0737$ & $\mathbf{0.0487 \pm 0.0381}$ & $\mathbf{0.1212 \pm 0.0572}$ \\
    \end{tabular}
    \label{tab:moveto}
\end{table}

\hfill \break
\noindent \textbf{Reaching preferred viewpoints.} Next, we consider an active agent that is tasked to reach a preferred observation that was provided in advance. To do so, the agent uses the generative model to encode both the preferred and initial observation and then uses Monte Carlo sampling to evaluate the expected free energy for 10000 potential actions, after which the action with the lowest expected free energy is executed. The expected free energy formulation is computed as the negative log probability of the latent representation with respect to the distribution over the preferred state, acquired through encoding the preferred observation. This is similar to the setup adopted by Van de Maele et al. \cite{vandemaele_2022}, with the important difference that now the preferred observation is an image of a \textit{different} object instance. 

\begin{figure}[t!]
    \centering
    \subfloat[]{\includegraphics[width=0.45\textwidth]{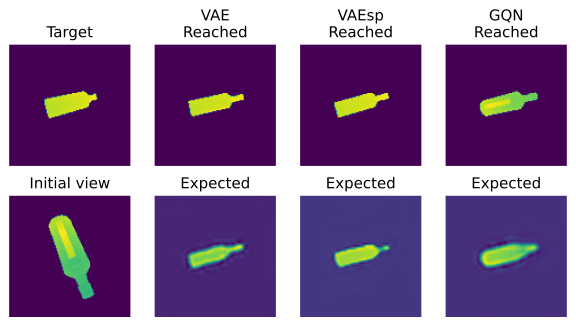}}
    \hfill
    \subfloat[]{\includegraphics[width=0.45\textwidth]{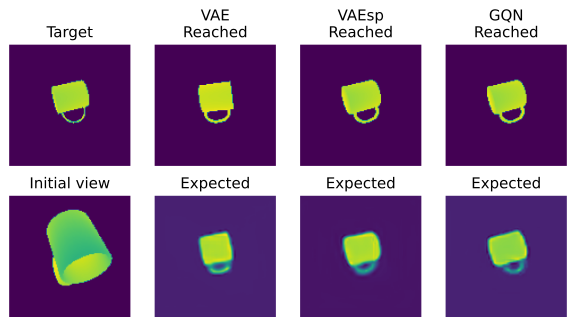}}
    \caption{Two examples of the experiment on reaching preferred viewpoints for a `bottle' (a) and a `mug' (b). First column shows the target view (top) and initial view given to the agent (bottom). Next, for the three models we show the actual reached view (top), versus the imagined expected view of the model (bottom).}
    \label{fig:movetoqual}
\end{figure}

To evaluate the performance, we compute the pixel-wise mean squared error (MSE) between a render of the target object in the preferred pose, and the render of the environment after executing the chosen action after the initial observation.
The results are shown in Table~\ref{tab:moveto}. VAEsp performs on par with the other approaches for `bowl' and `mug', but significantly outperforms the GQN on the the `bottle' and `can' categories, reflected by p-values of $0.009$ and $0.001$ for these respective objects. The p-values for the comparison with the VAE are $0.167$ and $0.220$, which are not significant. A qualitative evaluation is shown in Figure~\ref{fig:movetoqual}. Here we show the preferred target view, the initial view of the environment, as well as the final views reached by each of the agents, as well as what each model was imagining. Despite the target view being from a different object instance, the agent is able to find a matching viewpoint.

\begin{figure}[t!]
    \centering
    \subfloat[VAEsp bottle]{\includegraphics[width=4in]{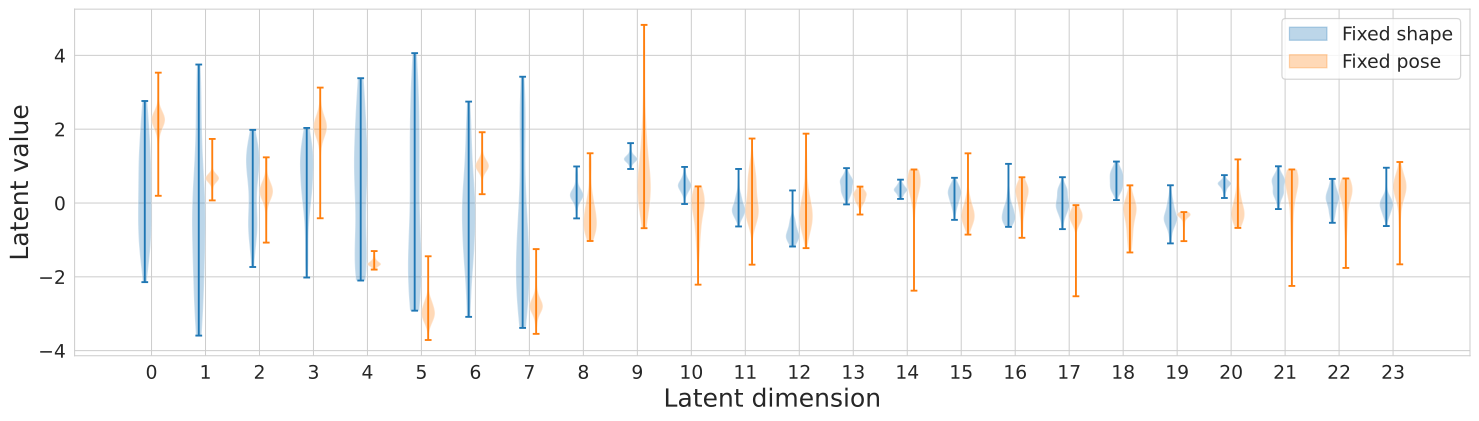}}
    \\
    \subfloat[VAEsp mug]{\includegraphics[width=4in]{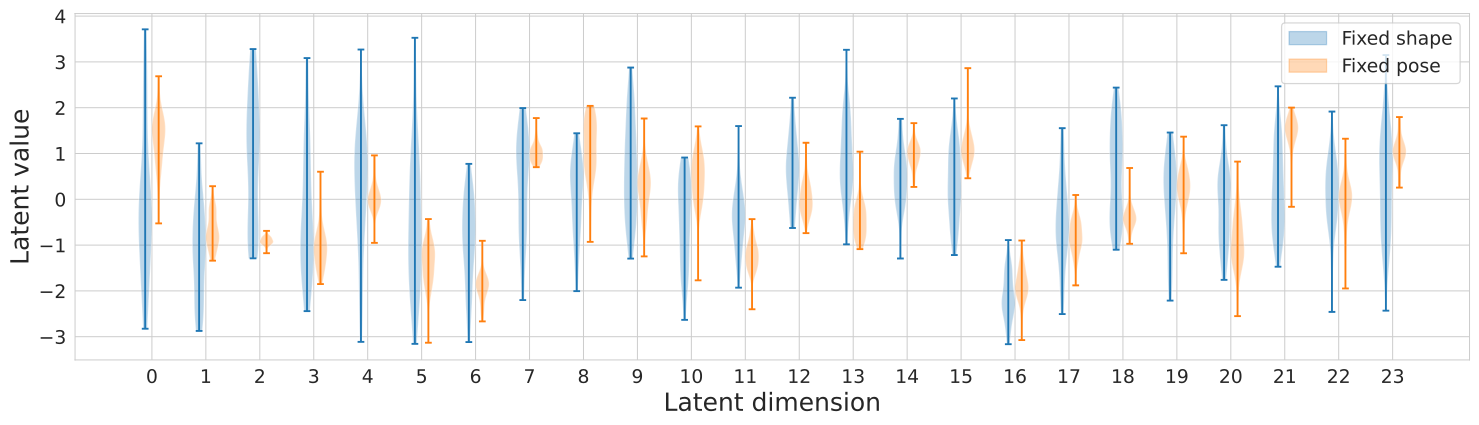}}
    \caption{Violin plots representing the distribution over the latent dimension when keeping either the pose or shape fixed. For the bottle model (a) the pose latent dimensions (0-7) vary when only varying the pose, whereas the shape latent dimensions (8-23) don't vary with the pose. For the mug model (b) we see the shape and pose latent are much more entangled.}
    \label{fig:violin}
\end{figure}

\hfill \break
\noindent \textbf{Disentangled latent space.}
Finally, we evaluate the disentanglement of shape and pose for the proposed architecture. Given that our VAEsp model outperforms the other models on `bottle` and `can`, but not on `bowl` and `mug`, we hypothesize that our model is able to better disentangle shape and pose for the first categories, but not for the latter. To evaluate this, we plot the distribution of each latent dimension when encoding 50 random shapes in a fixed pose, versus 50 random poses for a fixed shape, as shown on Figure~\ref{fig:violin}. We see that indeed the VAEsp model has a much more disentangled latent space for `bottle` compared to `mug`, which supports our hypothesis. Hence, it will be interesting to further experiment to find a correlation between latent space disentanglement and model performance. Moreover, we could work on even better enforcing disentanglement when training a VAEsp model, for example by adding additional regularization losses~\cite{factorvae,tcvae}. Also note that the GQN does not outperform the other models, although this one has access to the ground truth pose factor. This might be due to the fact that an SO(3) representation of pose is not optimal for the model to process, and it still encodes (entangled) pose information in the resulting latent space, as illustrated by violin plots for GQN models in Appendix~\ref{appendix:violins}.
Figure~\ref{fig:pose_shape} qualitatively illustrates the shape and pose disentanglement for our best performing model (bottle). We plot reconstructions of latent codes consisting of the shape latent of the first column, combined with the pose latent of the first row. 

\begin{figure}[t!]
  \centering
  \includegraphics[width=4.2in]{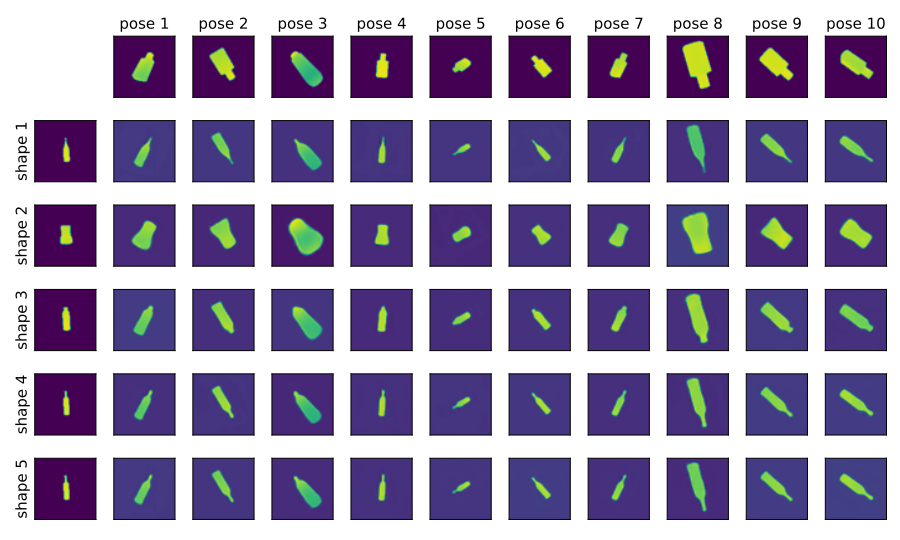}
  \caption{Qualitative experimentation for the bottle category. Images are reconstructed from the different pairings of the pose latent and shape latent of the first row and column respectively.}
  \label{fig:pose_shape}
\end{figure}

\section{Conclusion} 

In this paper, we proposed a novel deep active inference model for learning object-centric representations of object categories. In particular, we encourage the model to have a disentangled pose and shape latent code. We show that the better our model disentangles shape and pose, the better the results are on prediction, reconstruction as well as action selection towards a preferred observation.
As future work, we will further our study on the impact of disentanglement, and how to better enforce disentanglement in our model. We believe that this line of work is important for robotic manipulation tasks, i.e. where a robot learns to pick up a cup by the handle, and can then generalize to pick up any cup by reaching to the handle.

% uncomment for camera ready
%\subsubsection*{Acknowledgments}
%This research received funding from the Flemish Government under the ``Onderzoeksprogramma Artifici\"{e}le Intelligentie (AI) Vlaanderen'' programme.

\bibliographystyle{splncs04}
\bibliography{references}

\clearpage
\appendix 

%\section{The generative model}
%\label{appendix:generativemodel}

% derive free energy / expected free energy and loss function?

\section{Model and training details}
\label{appendix:details}

This paper compares three generative models for representing the shape and pose of an object. Each of the models has a latent distribution of $24$ dimensions, parameterized as a Gaussian distribution and has a similar amount of total trainable parameters.

\textbf{VAE}: The VAE baseline is a traditional variational autoencoder. The encoder consists of 6 convolutional layers with a kernel size of 3, a stride of 2 and padding of 1. The features for each layer are doubled every time, starting with 4 for the first layer. After each convolution, a LeakyReLU activation function is applied to the data. Finally, two linear layers are used on the flattened output from the convolutional pipeline, to directly predict the mean and log variance of the latent distribution. The decoder architecture is a mirrored version of the encoder. It consists of 6 convolutional layers with kernel size 3, padding 1 and stride 1. The layers have 32, 8, 16, 32 and 64 output features respectively. After each layer the LeakyReLU activation function is applied. The data is doubled in spatial resolution before each such layer through bi-linear upsampling, yielding a 120 by 120 image as final output. A transition model is used to predict the expected latent after applying an action. This model is parameterized through a fully connected neural network, consisting of three linear layers, where the output features are 64, 128 and 128 respectively. The input is the concatenation of a latent sample, and a 7D representation of the action (coordinate and orientation quaternion). The output of this layer is then again through two linear layers transformed in the predicted mean and log variance of the latent distribution. This model has $474.737$ trainable parameters.

\textbf{GQN}: The GQN baseline only consists of an encoder and a decoder. As the model is conditioned on the absolute pose of the next viewpoint, there is no need for a transition model. The encoder is parameterized exactly the same as the encoder of the VAE baseline. The decoder is now conditioned on both a latent sample and the 7D representation of the absolute viewpoint (coordinate and orientation quaternion). These are first concatenated and transformed through a linear layer with 128 output features. This is then used as a latent code for the decoder, which is parameterized the same as the decoder used in the VAE baseline. In total, the GQN has $361.281$ trainable parameters. 

\textbf{VAEsp}: Similar to the VAE baseline, the VAEsp consists of an encoder, decoder and transition model. The encoder is also a convolutional neural network, parameterized the same as the encoder of the VAE, except that instead of two linear layers predicting the parameters of the latent distribution, this model contains 4 linear layers. Two linear layers with 16 output features are used to predict the mean and log variance of the shape latent distribution, and two linear layers with 8 output features are used to predict the mean and log variance of the pose latent distribution. In the decoder, a sample from the pose and shape latent distributions are concatenated and decoded through a convolutional neural network, parameterized exactly the same as the decoder from the VAE baseline. The transition model, only transitions the pose latent, as we make the assumption that the object shape does not change over time. The transition model is parameterized the same as the transition model of the VAE, with the exception that the input is the concatenation of the 8D pose latent vector and the 7D action, in contrast to the 24D latent in the VAE. The VAEsp model has $464.449$ trainable parameters.

All models are trained using a constrained loss, where Lagrangian optimizers are used to weigh the separate terms~\cite{rezende_taming_2018}. During training, we tuned the reconstruction tolerance for each object empirically. Respectively to 'bottle', 'bowl', 'can' and 'mug' categories, MSE tolerances are: 350, 250, 280 and 520. Regularization terms are considered for each latent element. For all models, the Adam optimizer was used to minimize the objective.

\section{Additional qualitative results}
\label{appendix:qual}

\begin{figure}%
\centering
\subfloat[bottle]{\includegraphics[width=.5\textwidth]{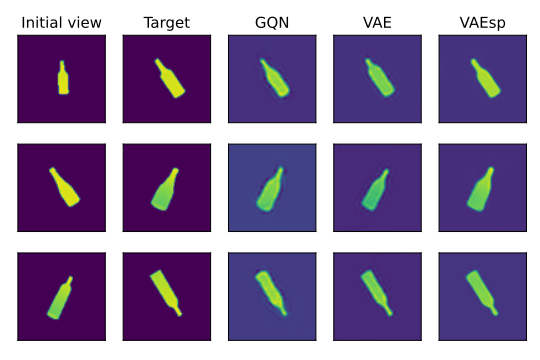}}
\subfloat[bowl]{\includegraphics[width=.5\textwidth]{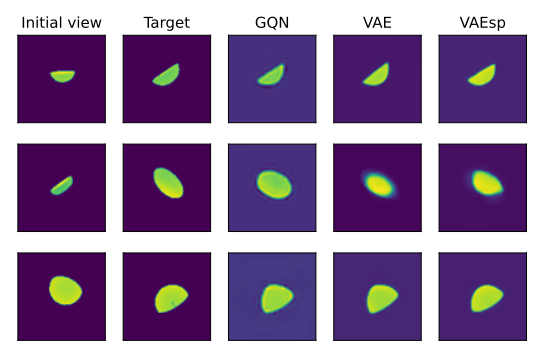}} \\
\subfloat[can]{\includegraphics[width=.5\textwidth]{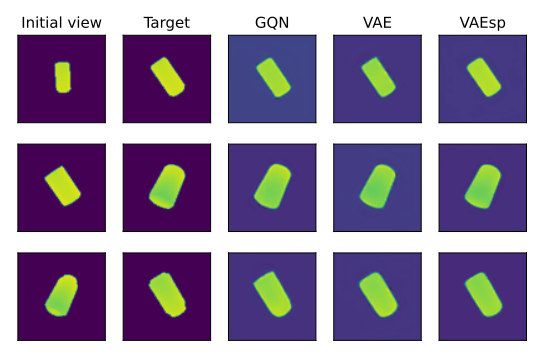}}
\subfloat[mug]{\includegraphics[width=.5\textwidth]{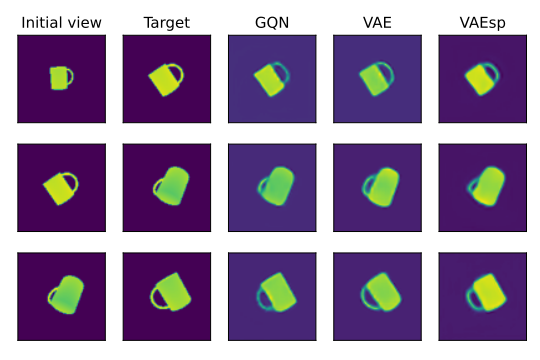}}
\caption{One-step prediction for different object categories.}
\label{fig:qualitative}
\end{figure}

\section{Latent disentanglement}
\label{appendix:violins}

\begin{figure}[t]
    \centering
    \subfloat[VAE bottle]{\includegraphics[width=\textwidth]{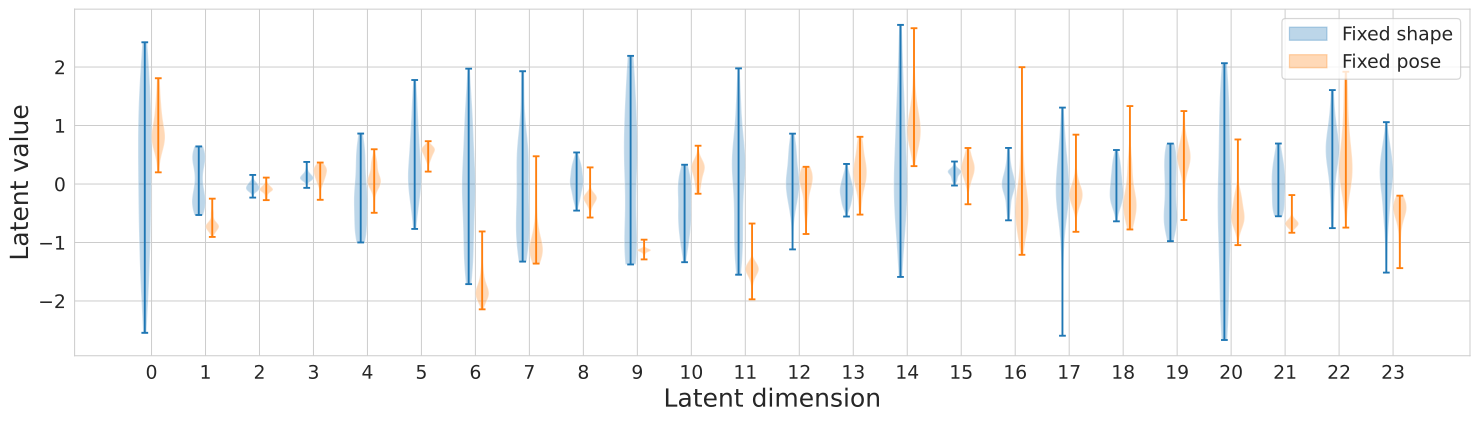}}
    \\
    \subfloat[GQN bottle]{\includegraphics[width=\textwidth]{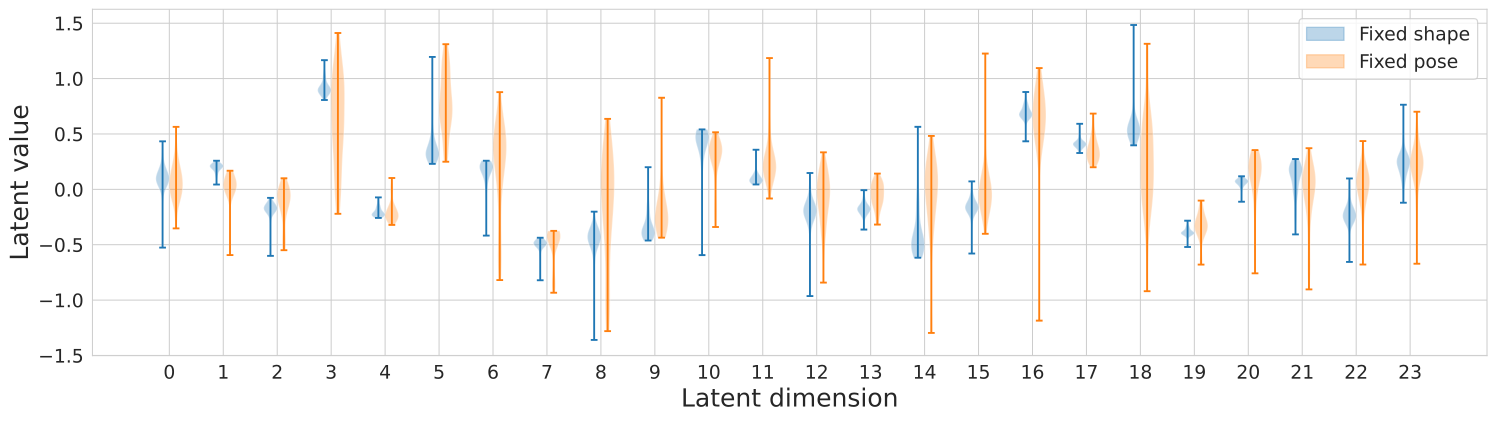}}
    \\
    \subfloat[VAEsp bottle]{\includegraphics[width=\textwidth]{figures/violin/VAEsp_bottle-violin.png}}
    \caption{Distribution of the latent values for the different models (VAE, GQN and VAEsp) for objects from the ``bottle'' class. In this experiment, 50 renders from a fixed object shape with a varying pose (fixed shape, marked in blue) are encoded. The orange violin plots represent the distribution over the latent values for 50 renders from the same object pose, with a varying object shape.}
    \label{fig:avbottle}
\end{figure}

\begin{figure}[t]
    \centering
    \subfloat[VAE can]{\includegraphics[width=\textwidth]{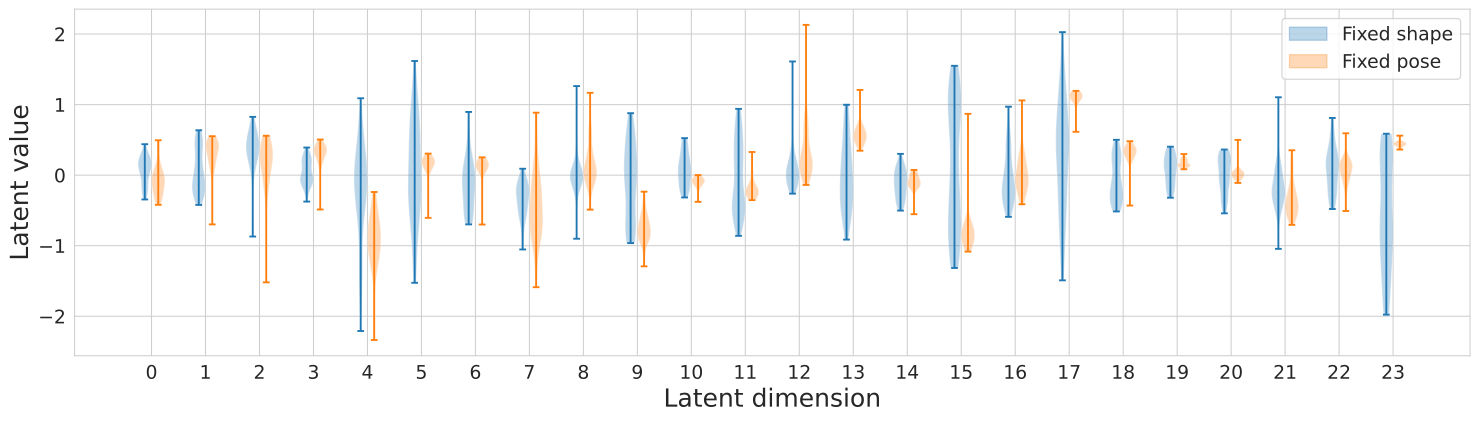}}
    \\
    \subfloat[GQN can]{\includegraphics[width=\textwidth]{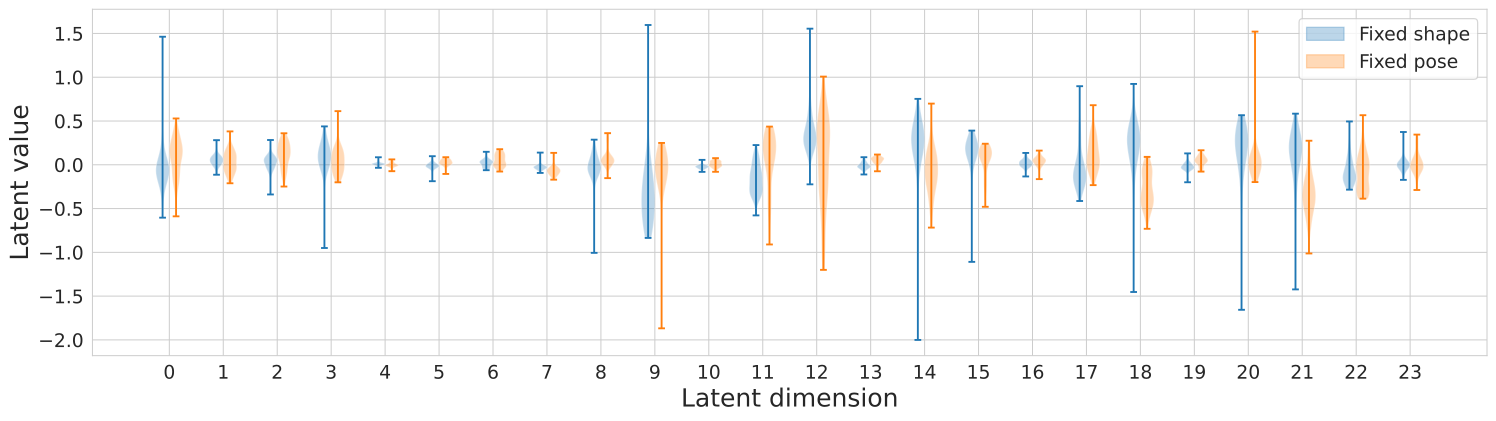}}
    \\
    \subfloat[VAEsp can]{\includegraphics[width=\textwidth]{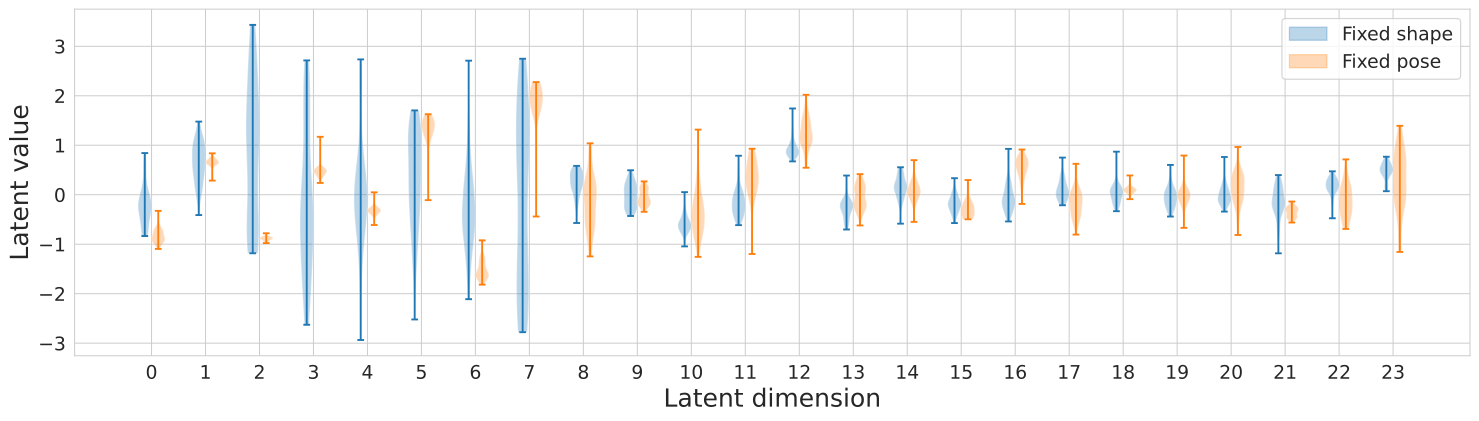}}
    \caption{Distribution of the latent values for the different models (VAE, GQN and VAEsp) for objects from the ``can'' class. In this experiment, 50 renders from a fixed object shape with a varying pose (fixed shape, marked in blue) are encoded. The orange violin plots represent the distribution over the latent values for 50 renders from the same object pose, with a varying object shape.}
    \label{fig:avcan}
\end{figure}

\begin{figure}[t]
    \centering
    \subfloat[VAE mug]{\includegraphics[width=\textwidth]{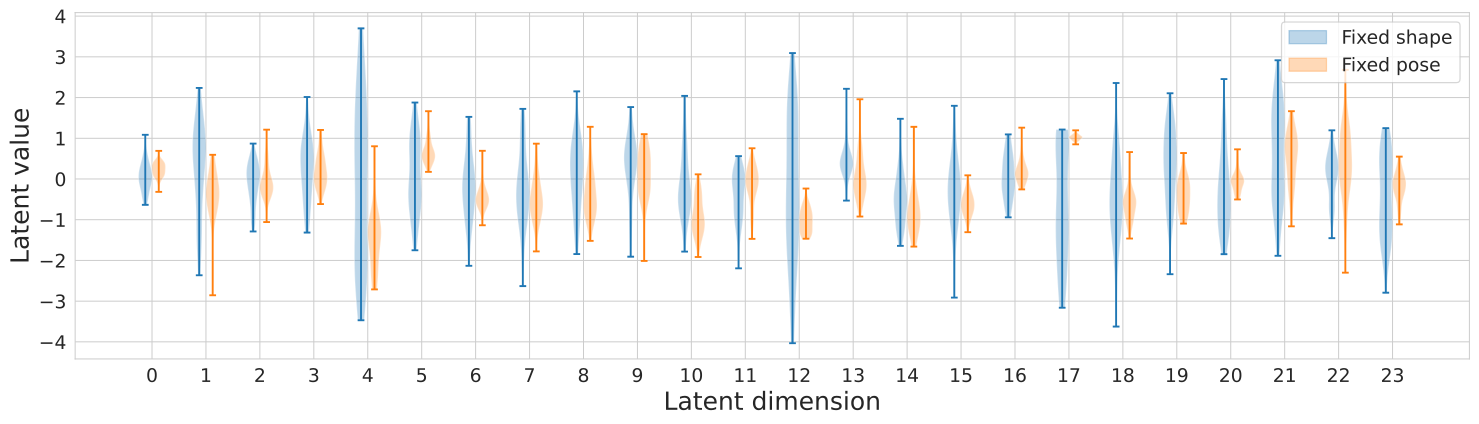}}
    \\
    \subfloat[GQN mug]{\includegraphics[width=\textwidth]{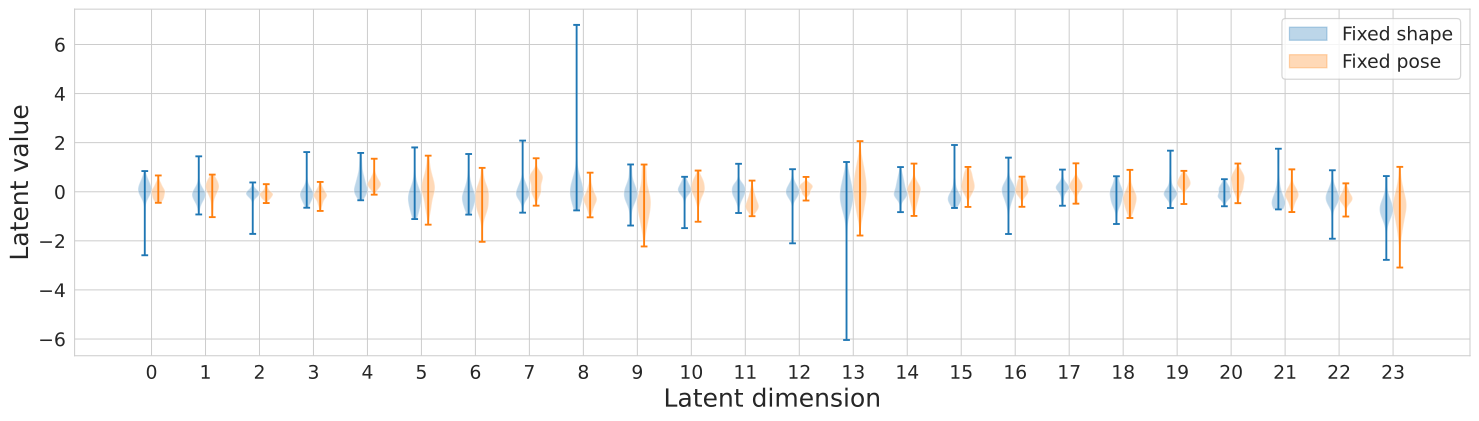}}
    \\
    \subfloat[VAEsp mug]{\includegraphics[width=\textwidth]{figures/violin/VAEsp_mug-violin.png}}
    \caption{Distribution of the latent values for the different models (VAE, GQN and VAEsp) for objects from the ``mug'' class. In this experiment, 50 renders from a fixed object shape with a varying pose (fixed shape, marked in blue) are encoded. The orange violin plots represent the distribution over the latent values for 50 renders from the same object pose, with a varying object shape.}
    \label{fig:avmug}
\end{figure}

\begin{figure}[t]
    \centering
    \subfloat[VAE bowl]{\includegraphics[width=\textwidth]{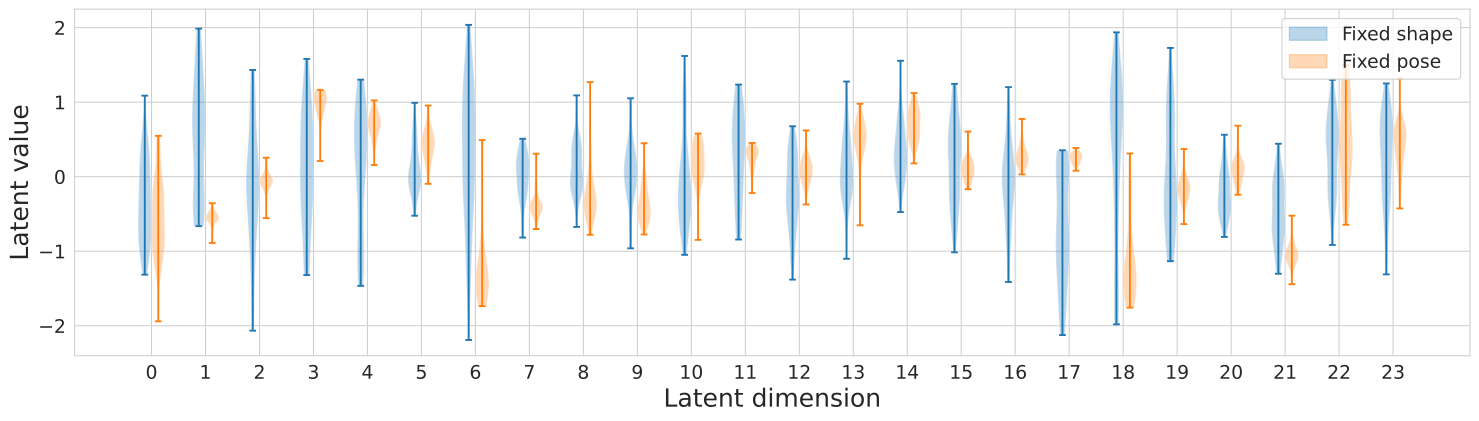}}
    \\
    \subfloat[GQN bowl]{\includegraphics[width=\textwidth]{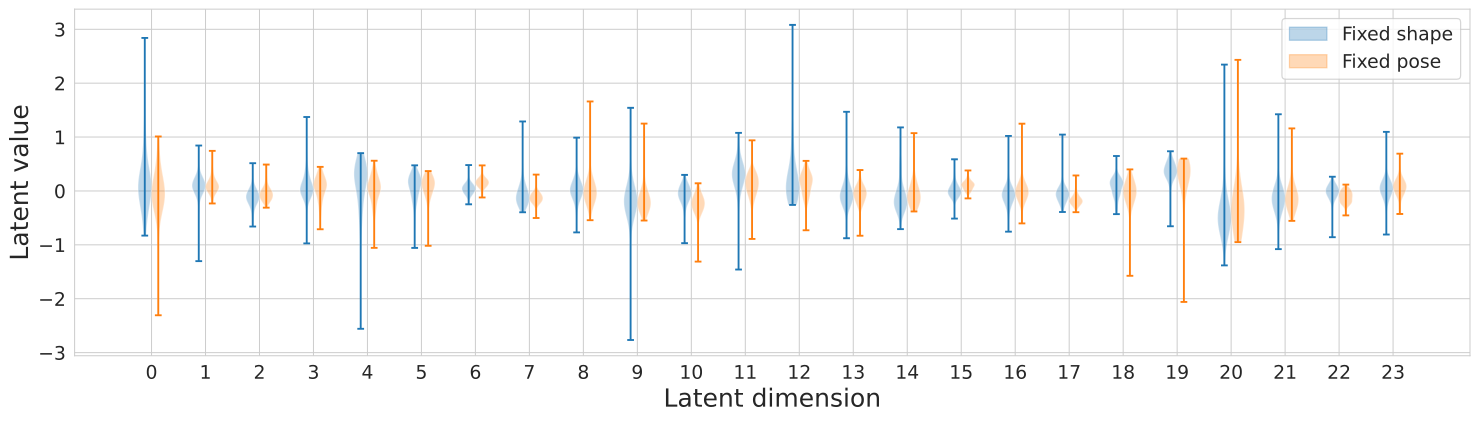}}
    \\
    \subfloat[VAEsp bowl]{\includegraphics[width=\textwidth]{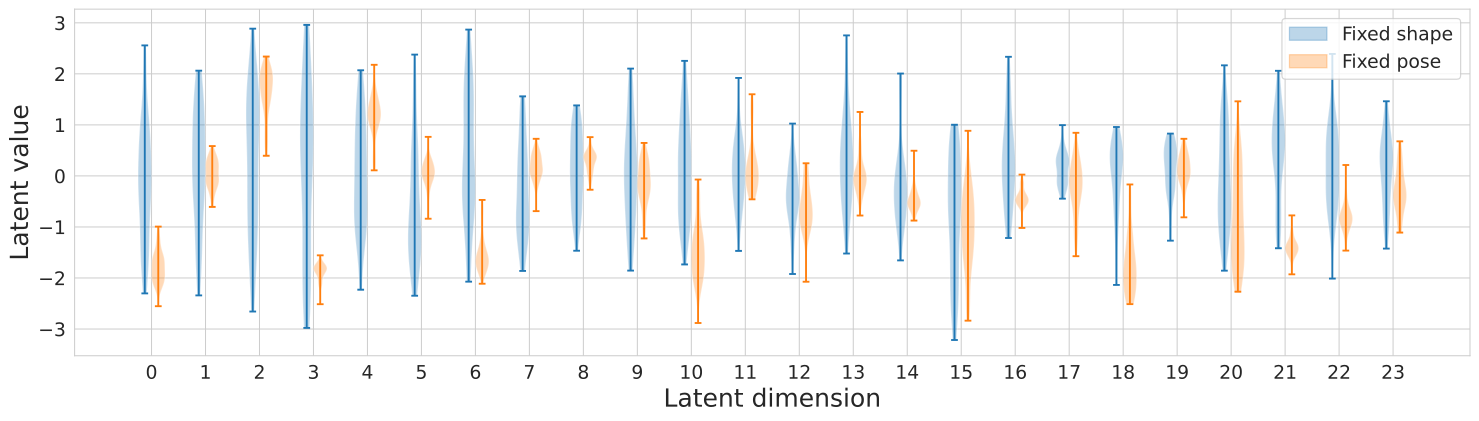}}
    \caption{Distribution of the latent values for the different models (VAE, GQN and VAEsp) for objects from the ``bowl'' class. In this experiment, 50 renders from a fixed object shape with a varying pose (fixed shape, marked in blue) are encoded. The orange violin plots represent the distribution over the latent values for 50 renders from the same object pose, with a varying object shape.}
    \label{fig:avbowl}
\end{figure}

In Figures~\ref{fig:avbottle},~\ref{fig:avcan},~\ref{fig:avmug} and~\ref{fig:avbowl}, we show the distribution over the latent values when encoding observation where a single input feature changes. The blue violin plots represent the distribution over the latent values for observations where the shape is kept fixed, and renders from different poses are fed through the encoder. The orange violin plots represent the distribution over the latent values for observations where the pose is kept fixed, and renders from different shapes within the object class are encoded through the encoder models. 

In these figures, we can clearly see that the encoding learnt by the VAE is not disentangled for any of the objects as the latent dimensions vary for both the fixed shape and pose cases. With the GQN, we would expect that the latent dimensions would remain static for the fixed shape case, as the pose is an explicit external signal for the decoder, however we can see that for a fixed shape, the variation over the latent value still varies a lot, in similar fashion as for the fixed pose. We conclude that the encoding of the GQN is also not disentangled. For the VAEsp model, we can see that in Figures~\ref{fig:avbottle} and~\ref{fig:avcan}, the first eight dimensions are used for the encoding of the pose, as the orange violins are much denser distributed for the fixed pose case. However, in Figures~\ref{fig:avmug} and~\ref{fig:avbowl}, we see that the model still shows a lot of variety for the latent codes describing the non-varying feature of the input. This result also strokes with our other experiments where for these objects both reconstruction as well as the move to perform worse.  

In this paper, we investigated the disentanglement for the different considered object classes. We see that our approach does not yield a disentangled representation each time. Further investigation and research will focus on better enforcing this disentanglement. 

\end{document}